\documentclass[lettersize,journal]{IEEEtran}
\usepackage{amsmath,amsfonts}
\usepackage{algorithmic}
\usepackage{algorithm}
\usepackage{array}
\usepackage[caption=false,font=normalsize,labelfont=sf,textfont=sf]{subfig}
\usepackage{textcomp}
\usepackage{stfloats}
\usepackage{url}
\usepackage{verbatim}
\usepackage{graphicx}
\usepackage{makecell}
\usepackage{ulem}
\usepackage{multirow}
\usepackage{booktabs}
\usepackage{threeparttable}
\usepackage{cite}
\begin{document}

\title{GMoE-FTI: A Universal Facial Template Inversion Method Adopting Genuine Mixture-of-Experts Mechanism}

\author{Zixuan Shen, Zhihua Xia, Kaikai Gan, Peipeng Yu
        % <-this % stops a space

\thanks{\textit{(Corresponding author: Zhihua Xia.)}}% <-this % stops a space
\thanks{Zixuan Shen, Zhihua Xia, and Kaikai Gan are with the College of Cyber Security,  Jinan University,  Guangzhou, Guangdong 511443, China (e-mail:
m15195901120@163.com; xia\_zhihua@163.com; kkgan406@163.com).}
\thanks{Peipeng Yu is with the School of Cyber Science and Engineering,  Nanjing University of Information Science \& Technology,  Nanjing, Jiangsu 210044, China (e-mail:
ypp865@163.com).}}

% The paper headers
%\markboth{Journal of \LaTeX\ Class Files,~Vol.~14, No.~8, August~2021}%
%{Shell \MakeLowercase{\textit{et al.}}: A Sample Article Using IEEEtran.cls for IEEE Journals}

%\IEEEpubid{0000--0000/00\$00.00~\copyright~2021 IEEE}
% Remember, if you use this you must call \IEEEpubidadjcol in the second
% column for its text to clear the IEEEpubid mark.

\maketitle

\begin{abstract}
Face recognition systems have been ubiquitously deployed across daily scenarios. To protect user privacy, mainstream systems store facial templates rather than raw face images for identity verification. However, facial template inversion (FTI) technologies can reconstruct faces from templates, which endangers biometric privacy. Existing FTI methods only fit single type of templates, failing to reconstruct identity-consistent faces from unseen types. To tackle this limitation, we survey mainstream template extractors which can be classified into five categories according the template distribution: unit-sphere, angle-linear, strict unit-sphere, cone-shaped, and uniform hypersphere. Inspired by this observation, this paper proposes GMoE-FTI, a \textbf{universal} FTI framework powered by a genuine MoE mechanism that differs from load-balanced multi-branch alternatives. It follows a three-stage training strategy: (1) independently train five mapping experts (\textbf{ME}s) and a matching module (\textbf{MM}); (2) conduct collaborative inference by weighting each \textbf{ME}’s output with \textbf{MM}’s output, so as to generate unified latent codes compatible with pre-trained StyleGAN3; (3) perform end-to-end joint fine-tuning over all modules to identity consistency and boost cross-expert coordination. Extensive evaluations on the LFW benchmark validate the superiority of GMoE-FTI against SOTA methods. Evaluated on seven prevailing unseen template types including AdaFace, CosFace, CurricularFace, ElasticFace, SphereFace2, SphereFace-R, and TransFace, GMoE-FTI achieves an average 40\% TAR improvement, with average FAPD and FAPC each reduced by 2\%, verifying its universal generalization across heterogeneous templates. 
\end{abstract}


\section{Introduction}
\IEEEPARstart{D}{riven} by contact-free operation and convenient usage, face recognition (FR) systems have been extensively applied in daily life \cite{hasan2023presentation}. Stricter privacy regulations \cite{gdpr2016,ChinaFaceRecognitionMeasures2025} and rising public awareness of biometric security have driven mainstream FR systems to adopt facial templates \cite{huang2020curricularface} rather than raw face images \cite{bowyer2004face} for identity authentication. However, facial templates encode highly sensitive identity features. It has been demonstrated that facial template inversion (FTI) technologies \cite{dai2026clipfti,otroshi2023face} can reconstruct recognizable faces from leaked templates, leading to severe privacy leakage risks.

\begin{figure}[ht]
    \centering
    \includegraphics[width=\linewidth]{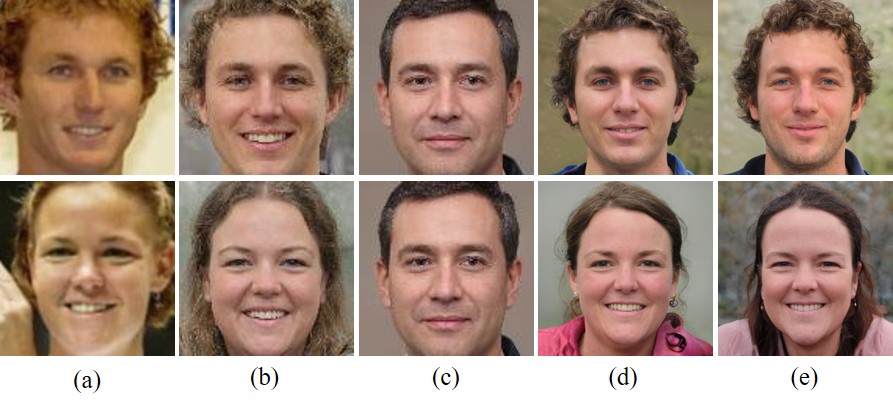}
    \vspace{-0.45cm}
    \caption{Limitations of existing FTI methods: (a) the original face images, (b) identity-consistent face images reconstructed from ArcFace templates by a representative SOTA FTI approach \cite{dai2026clipfti} trained by ArcFace templates, (c) identity-inconsistent face images reconstructed from AdaFace templates by \cite{dai2026clipfti}, (d) identity-consistent face images from ArcFace templates by our model trained by five types of templates, and (e) identity-consistent face images reconstructed from AdaFace templates by our model. Please note that, the training set of our model do \textbf{NOT} include AdaFace samples.     
    }
    \label{fig:Compare}
\end{figure}

Existing SOTA FTI models are optimized solely for a single template type during training. As illustrated in Fig. \ref{fig:Compare}, they achieve identity-consistent inversion only when processing templates derived from the FR model seen in training. When fed templates from unseen extractors, they reconstruct faces with prominent identity-inconsistency. In real attack scenarios, adversaries often have no prior knowledge of the source extractor that generated leaked templates; worse still, such templates may stem from unreleased extractor variants. Obviously, training a dedicated inversion model for every possible template extractor is computationally infeasible.

To alleviate this problem, we systematically investigate mainstream facial template extractors and propose GMoE-FTI, a universal FTI framework built on a genuine mixture-of-experts (MoE) architecture paired with a three-stage training pipeline. Our GMoE-FTI framework achieves identity-consistent inversion across heterogeneous facial templates sourced from a range of extractors. The contributions of this work can be summarized as follows:

\begin{itemize}
    \item \textbf{A investigation on facial template distributions}: We investigate the mainstream facial template extractors proposed from 2015 to the present. It is revealed that the loss functions in the training of these prevailing are predominantly built upon Softmax and its variants, which regularize facial templates into five distinct categories according to the template distributions: unit-sphere, angle-linear, strict unit-sphere, cone-shaped, and uniform hypersphere. Based on this taxonomy, we trained our universal inverter with training samples from a compact set of template extractors.

    \item \textbf{A genuine MoE paradigm tailored for FTI}: Existing MoE architectures adopt multi-branch structures mainly for load balancing and channel splitting. All branches are trained jointly without specialized expertise for distinct data manifolds. To accommodate heterogeneous facial template distributions for universal FTI, we also propose multi-branch structure named genuine MoE (GMoE). It consists of five independent expert branches, each specialized for one unique template distribution. Supported by a customized matching modules, GMoE feeds arbitrary input templates to all expert branches and aggregates multi-branch outputs via weighted fusion to produce unified face reconstruction.

    \item \textbf{Universal FTI capability}: GMoE-FTI establishes a universal FTI paradigm to alleviate the single-template-domain restriction of existing FTI methods, compatible with arbitrary facial templates of varying dimensions from unknown extractors. Extensive evaluations show our method surpasses four mainstream SOTA FTI approaches: Adv-StyleGAN3 (NeurIPS 2023), MSE-StyleGAN3 (IJCB 2023), DSCasConv (TIFS 2024) and CLIP-FTI (AAAI 2026). Specifically, GMoE-FTI delivers an average 40\% TAR gain alongside 2\% reductions in both FAPD and FAPC.
\end{itemize}

\section{Related Work}

\subsection{Facial Template Inversion}

Existing facial template inversion (FTI) research can be split into three developmental stages: handcrafted feature inversion \cite{adler2003sample,mignon2013reconstructing}, confidence vector inversion \cite{fredrikson2015model}, and deep facial feature inversion \cite{mai2018reconstruction}. This work focuses on the third stage, as mainstream face recognition systems adopt deep facial features as storage templates. Relevant methods are further split into two technical routes: custom-built inversion and pre-trained network-based inversion.

\textbf{Custom-built network-based FTI}. Methods in this category build end-to-end template-to-image mappings via stacked deconvolutional and convolutional blocks. Mai et al. \cite{mai2018reconstruction} and Shahreza et al. \cite{shahreza2022face,shahreza2024vulnerability} stacked convolution–deconvolution blocks equipped with dense skip connections and multi-kernel parallel convolutions, to enhance fine-grained detail reconstruction and identity preservation. Furthermore, Duong et al. \cite{duong2020vec2face} proposed the feature-conditioned generator, which injects facial templates as conditional inputs into multi-scale feature maps of the generator. The model is jointly optimized via pixel loss, perceptual loss, and bijection metric loss to improve reconstruction fidelity and identity consistency.

\textbf{Pre-trained network-based FTI}. With the widespread adoption of StyleGAN for high-resolution face generation, most modern FTI methods adopt pre-trained generative networks as backbones and learn mappings between facial templates and latent codes of the generator. Early work by Dong et al. \cite{dong2021towards} trained regression models to search optimal latent codes for StyleGAN2. Subsequent works uses genetic algorithms \cite{dong2023reconstruct} and simulated annealing \cite{vendrow2021realistic} to refine latent code search and alleviate regression overfitting. Recent studies \cite{otroshi2023face,shahreza2023inversion,shahreza2024template} design dedicated mapping networks to project templates into intermediate latent spaces of StyleGAN3, optimized via latent discriminator loss, pixel loss, and identity loss. Further extensions integrate attribute alignment adapters \cite{0Attribute} and CLIP semantic embeddings \cite{dai2026clipfti} to strengthen attribute consistency of reconstructed images. Futhermore, Shahreza et al. \cite{huang2023comprehensive,shahreza2023template} extended this paradigm to 3D face reconstruction via pre-trained GNeRF generators embedding StyleGAN2 cores. Moreover, Shen et al. \cite{Shen2026SWFTIFT} designed latent-level and image-level regularization losses for StyleSwin inversion. Shahreza et al. \cite{Shahreza_2025_CVPR} proposed an adapter module to project templates into Arc2Face embedding space, then inject the transformed features into the UNet decoder of pre-trained Stable Diffusion for face reconstruction.

Overall, as shown in Table \ref{tab:related_work}, 
although existing methods can achieve identity-consistent face reconstruction, they suffer from prominent practical drawbacks: they require prior knowledge of the extractor generating target templates, and the inversion network is only optimized for that single extractor. However, in real attack scenarios, the source of leaked templates is often unknown, highlighting the demand for generalizable inverters compatible with arbitrary unknown facial templates.

\begin{table*}[htbp]
\centering
\caption{Related research on facial template inversion}
\small
\label{tab:related_work}
\begin{tabular}{p{2.0cm} p{2.0cm} p{4.0cm} p{4.5cm} p{1.8cm} p{1.2cm}}
\toprule
\textbf{References} & \textbf{Types} & \textbf{Inverter Backbones} & \textbf{Target Models} & \textbf{Universal} & \textbf{Code} \\
\midrule
\cite{mai2018reconstruction} & \multirow{4}{*}{Custom-Built} & NbNet (Deconv+Conv) & FaceNet & $\times$ & $\checkmark$ \\
\cite{duong2020vec2face} & & FCG & ArcFace & $\times$ & $\times$ \\
\cite{shahreza2022face} & & Deconv+Conv & ArcFace, ElasticFace & $\times$ & $\checkmark$ \\
\cite{shahreza2024vulnerability} & & DSCasConv (Deconv+Conv) & ArcFace & $\times$ & $\checkmark$ \\
\midrule
\cite{dong2021towards} & \multirow{12}{*}{\makecell[l]{Pre-trained \\Network-based}} & StyleGAN2 & ArcFace & $\times$ & $\times$ \\
\cite{vendrow2021realistic} & & StyleGAN2 & FaceNet & $\times$ & $\checkmark$ \\
\cite{dong2023reconstruct} & & StyleGAN2 & ArcFace & $\times$ & $\checkmark$ \\
\cite{shahreza2023inversion} & & StyleGAN3 & ArcFace, ElasticFace & $\times$ & $\checkmark$ \\
\cite{otroshi2023face} & & StyleGAN3 & ArcFace, ElasticFace & $\times$ & $\checkmark$ \\
\cite{huang2023comprehensive} & & GNeRF (StyleGAN2) & ArcFace, ElasticFace & $\times$ & $\checkmark$ \\
\cite{shahreza2023template} & & GNeRF (StyleGAN2) & ArcFace, ElasticFace & $\times$ & $\checkmark$ \\
\cite{shahreza2024template} & & StyleGAN3 & ArcFace, ElasticFace & $\times$ & $\checkmark$ \\
\cite{Shahreza_2025_CVPR} & & Stable Diffusion & ArcFace, ElasticFace & $\times$ & $\times$ \\
\cite{dai2026clipfti} & & CLIP-FTI (StyleGAN3) & ArcFace, ElasticFace & $\times$ & $\checkmark$ \\
\cite{0Attribute} &  & AMA-CLN (StyleGAN3) & ArcFace, ElasticFace & $\times$ & $\checkmark$ \\
\cite{Shen2026SWFTIFT} & & StyleSwin & ArcFace, SwinFace & $\times$ & $\checkmark$ \\
\midrule 
& \multirow{2}{*}{Ours} & \multirow{2}{*}{Our GMoE-FTI (StyleGAN3)} & FaceNet, SphereFace, ArcFace, MagFace, SphereFace+ & \multirow{2}{*}{$\checkmark$} & \multirow{2}{*}{$\checkmark$} \\
\bottomrule
\end{tabular}
\end{table*}

\subsection{Existing Mixture-of-Experts Mechanism}

The Mixture-of-Experts (MoE) \cite{fedus2022switch} architecture uses modular multi-branch designs to expand model capacity and boost representation power for complex tasks, widely applied in large language models and computer vision generation pipelines. An MoE framework consists of a gating scheduler network and multiple expert branches. During training, the gating module dynamically activates partial sub-networks to balance computational overhead and model fitting capacity.

Nevertheless, existing MoE architectures possess intrinsic structural limitations when applied to universal FTI task: expert branches are randomly initialized and co-trained without dedicated pre-training priors, resulting in homogeneous branch performance with no specialized fitting ability for distinct feature distributions. Additionally, the gating module only execute superficial top-k expert selection and enforce batch-level load-balancing regularization, while lacking dedicated modules for feature distribution perception, distribution-aligned expert matching, and cooperative joint inference, making it unsuitable for universal FTI tasks. To our best knowledge, no prior FTI literature proposes a genuine MoE framework built upon independently pre-trained heterogeneous experts equipped with template-aware matching and collaborative inference modules.

\section{System and Security Model}

\subsection{Model Definition}

In face recognition (FR) pipelines, universal FTI tasks can be mathematically formalized with two core mapping functions:

\textbf{Facial template extractor:} Let $\mathcal{E}: \mathcal{X} \rightarrow \mathcal{T}$ denote the facial template extraction mapping. It projects an arbitrary face image $x \in \mathcal{X}$ into a $d$-dimensional facial template $t \in \mathcal{T}$. Current facial template extractors can be divided into different categories according to the distribution of their templates.

\textbf{Facial template inverter:} Let $\mathcal{G}: \mathcal{T} \rightarrow \mathcal{X}$ denote the universal FTI mapping. It reconstructs face images from templates. In our framework, the inverter is implemented as a generative network $g_{\theta}$ with learnable parameters $\theta$.

\subsection{Adversary' Capabilities}

We suppose the adversary can access facial template extractors in black-box manner, which is a prevalent paradigm in prior FTI works and aligned with real-world scenarios.

For each type of template extractor, the adversary gains black-box access to one model within the category. The adversary may input arbitrary face images to retrieve the corresponding templates $t = \mathcal{E}(x)$. With these outputs, the adversary constructs multi-category paired datasets $\{(x, t)\}$for training the universal inverter.

\subsection{Optimization Objectives}
% for Universal Inversion

Given an arbitrary target template $t^* = \mathcal{E}(x^*)$, we train the universal inverter $g_{\theta}$ under three mutually complementary optimization goals aligned with practical attack scenarios:

\textbf{Identity consistency for in-training templates:} When fed with templates from extractors seen during training, the reconstructed image $\hat{x} = \mathcal{G}(t^*)$ must be visually similar to the original image $x^*$. The template $\hat{t} = \mathcal{E}(\hat{x})$ extracted from the reconstruction needs high similarity with $t^*$, so the generated image can pass identity authentication of the source FR model and guarantee intra-category identity consistency.

\textbf{Identity consistency for unseen templates:} For the template $t'$ produced by the extractor excluded from training, the model still generates visually face image $\hat{x} = \mathcal{G}(t')$ similar to the original image $x'$, and the template extracted from the reconstruction maintains high similarity to $t'$.

\section{An investigation into Facial Template Extraction}

Current facial template extractors leverage loss functions to regularize heterogeneous template spaces. Softmax loss serves as the baseline for facial representation learning:
\begin{equation}
    L_{\text{Softmax}} = -\frac{1}{N} \sum_{i=1}^{N} \log \frac{e^{W_{y_i}^{T} x_i + b_{y_i}}}{\sum_{j=1}^{C} e^{W_{j}^{T} x_i + b_j}},
\end{equation}

\noindent where $x_i$ denotes input feature of image $i$, $y_i$ is the ground-truth identity label, $W$ and $b$ represent the weight and bias of fully-connected layer, $N$ is the number of training samples, and $C$ denotes the total identity classes. 

As Softmax lacks geometric regularization, it generates scattered templates with weak intra-class compactness, hurting recognition accuracy. Subsequent improved loss designs impose distinct geometric regularization on features ($x$) or/and weights ($W$), dividing facial templates into five categories according to the template distributions: unit-sphere, angle-linear, strict unit-sphere, cone-shaped, and uniform hypersphere. We elaborate each category below.

%To tackle this limitation, existing improvements mainly fall into two categories: feature/weight normalization and angular margin regularization.

\textbf{Unit-sphere Distribution}.
This category normalizes feature $x$ to map facial templates onto a \uline{unit hypersphere}. FaceNet \cite{schroff2015facenet} pioneered this scheme. However, unnormalized weights break spherical consistency and restrict inter-class discriminability.

\textbf{Angle-linear Distribution}. SphereFace (A-Softmax) \cite{liu2017sphereface} normalizes weight $W$ to impose \uline{linear angular} constraints on weights, so as to boost inter-class discriminability. Nevertheless, unnormalized features introduce magnitude bias and yield inferior intra-class clustering.

\textbf{Strict Unit-sphere Distribution}. Inspired by the two prior normalization schemes, NormFace \cite{wang2017normface} jointly normalizes features and weights to strengthen intra-class compactness and inter-class angular separation, with a learnable scale factor $s$ that widens identity angular gaps:
\begin{equation}
    L_{\text{NormFace}} = -\frac{1}{N} \sum_{i=1}^{N} \log \frac{e^{s \cdot \cos \theta_{y_i}}}{\sum_{j=1}^{C} e^{s \cdot \cos \theta_j}}.
\end{equation}

Dual normalization eliminates magnitude bias and projects embeddings onto a \uline{unified unit sphere}, laying a foundational baseline for facial template regularization. Building on this paradigm, CosFace \cite{wang2018cosface} and ArcFace \cite{deng2019arcface} further incorporate additive margins within the cosine and angular spaces, respectively, to enhance intra-class clustering while preserving stable training dynamics. CurricularFace \cite{huang2020curricularface} leverages curriculum learning to progressively raise the margin constraint difficulty. ElasticFace \cite{boutros2022elasticface}, X2-Softmax \cite{xu2024x2}, and JAMsFace \cite{khalifa2023jamsface} devise adaptive margin schemes under full spherical normalization. State-of-the-art Transformer-based extractors such as TransFace \cite{dan2023transface} and LVFace \cite{you2025lvface} merely replace CNN backbones yet still adopt rigid unit-sphere regularization as the core constraint.

\begin{table*}[ht]
\centering
\caption{Taxonomy of mainstream facial template extractors (2015–Present).}
\label{tab:templates}
\small
\begin{threeparttable}
\begin{tabular}{m{2.8cm}m{2.5cm}m{2.5cm}m{2.5cm}m{2.5cm}m{2.7cm}}
\toprule
\textbf{Feature Distribution} & Unit-sphere & Angle-linear & Strict unit-sphere & Cone-shaped & Uniform hypersphere \\
\midrule
\makecell[c]{\textbf{Illustrations}} & \includegraphics[width=2.0cm]{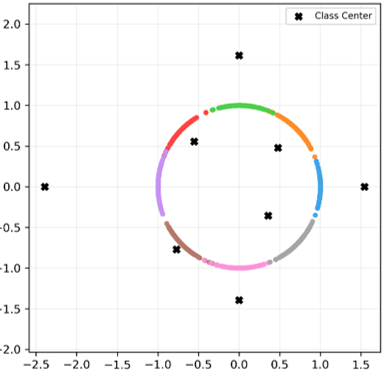} & \includegraphics[width=2.0cm]{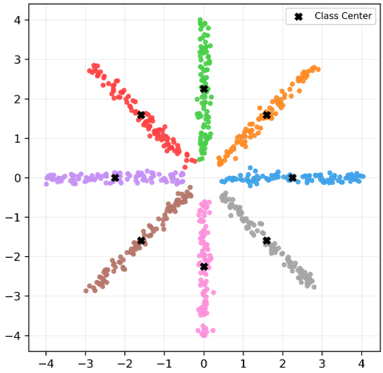} & \includegraphics[width=2.0cm]{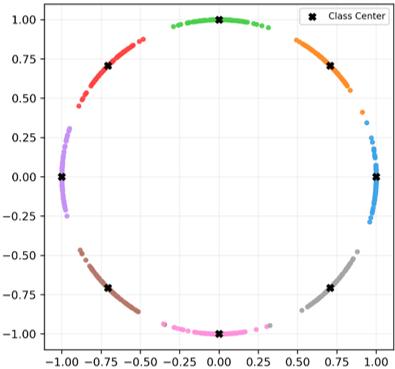} & \includegraphics[width=2.0cm]{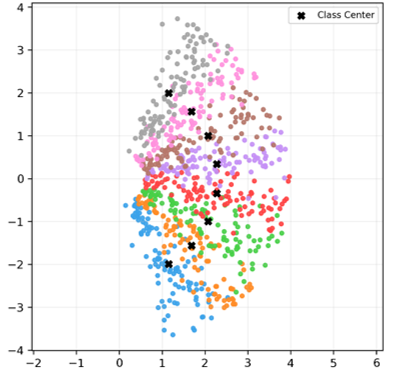} & \includegraphics[width=2.0cm]{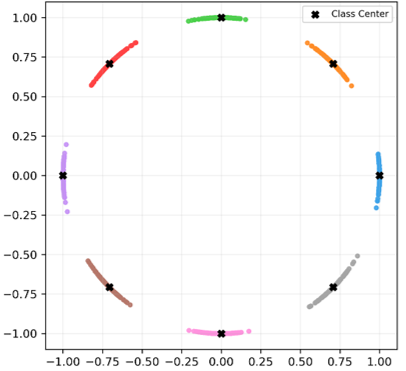} \\
\bottomrule
\end{tabular}
\begin{tablenotes}[flushleft]
\scriptsize
\item[†] Note: Unit-sphere contains FaceNet \cite{schroff2015facenet}; Angle-linear contains SphereFace (A-Softmax) \cite{liu2017sphereface}; Strict unit-sphere contains NormFace \cite{wang2017normface}, CosFace \cite{wang2018cosface}, ArcFace \cite{deng2019arcface}, CurricularFace \cite{huang2020curricularface}, ElasticFace \cite{boutros2022elasticface}, SphereFace2 \cite{wen2021sphereface2}, SphereFace-R \cite{liu2022sphereface}, X2-Softmax \cite{xu2024x2}, JAMsFace \cite{khalifa2023jamsface}, TransFace \cite{dan2023transface}, LVFace \cite{you2025lvface}; Cone-shaped contains MagFace \cite{meng2021magface}, AdaFace \cite{kim2022adaface}, KappaFace \cite{oinar2023kappaface}, HAMFace \cite{li2024hamface}, ExpFace \cite{zheng2025expface}; Uniform hypersphere contains SphereFace+ \cite{liu2018learning}.
\end{tablenotes}
\end{threeparttable}
\end{table*}

\textbf{Cone-shaped Distribution}. Mandatory unit-sphere projection treats high-quality and blurry/occluded low-quality faces identically, weakening discriminative power on hard samples and long-tailed data. To resolve this issue, MagFace \cite{meng2021magface} and AdaFace \cite{kim2022adaface} discard rigid unit normalization and link template norms to facial quality, yielding \uline{cone-shaped} manifolds where high-quality faces correspond to larger vector magnitudes. Subsequent follow-ups including KappaFace \cite{oinar2023kappaface}, HAMFace \cite{li2024hamface}, and ExpFace \cite{zheng2025expface} inherit this magnitude-aware modeling design and further refine adaptive regularization strategies based on the cone-shaped feature distribution.

\textbf{Uniform Hypersphere Distribution}. Built upon SphereFace, SphereFace+ (MHE) \cite{liu2018learning} incorporates hypersphere energy regularization to evenly scatter templates across the sphere, constructing the so-called \uline{uniform hypersphere} distribution. Conventional spherical losses cluster templates of frequent identities locally and create unbalanced spherical occupancy; this energy penalty alleviates the imbalance. Still, this paradigm incurs extra computational cost and achieves negligible accuracy gains over dominant strict unit-sphere losses, attracting minimal follow-up research.

As summarized in Table \ref{tab:templates}, we systematically survey mainstream facial template extractors from 2015 to the present and categorize them into five distinct geometric distributions: \uline{unit-sphere}, \uline{angle-linear}, \uline{strict unit-sphere}, \uline{cone-shaped}, and \uline{uniform hypersphere}. Inspired by this observation, we train a dedicated expert branch for each of the five representative extractors, forming the core of our collaborative genuine MoE inversion pipeline.

\section{Methodology} 
\begin{figure}[htbp]
    \centering
    \includegraphics[width=\linewidth]{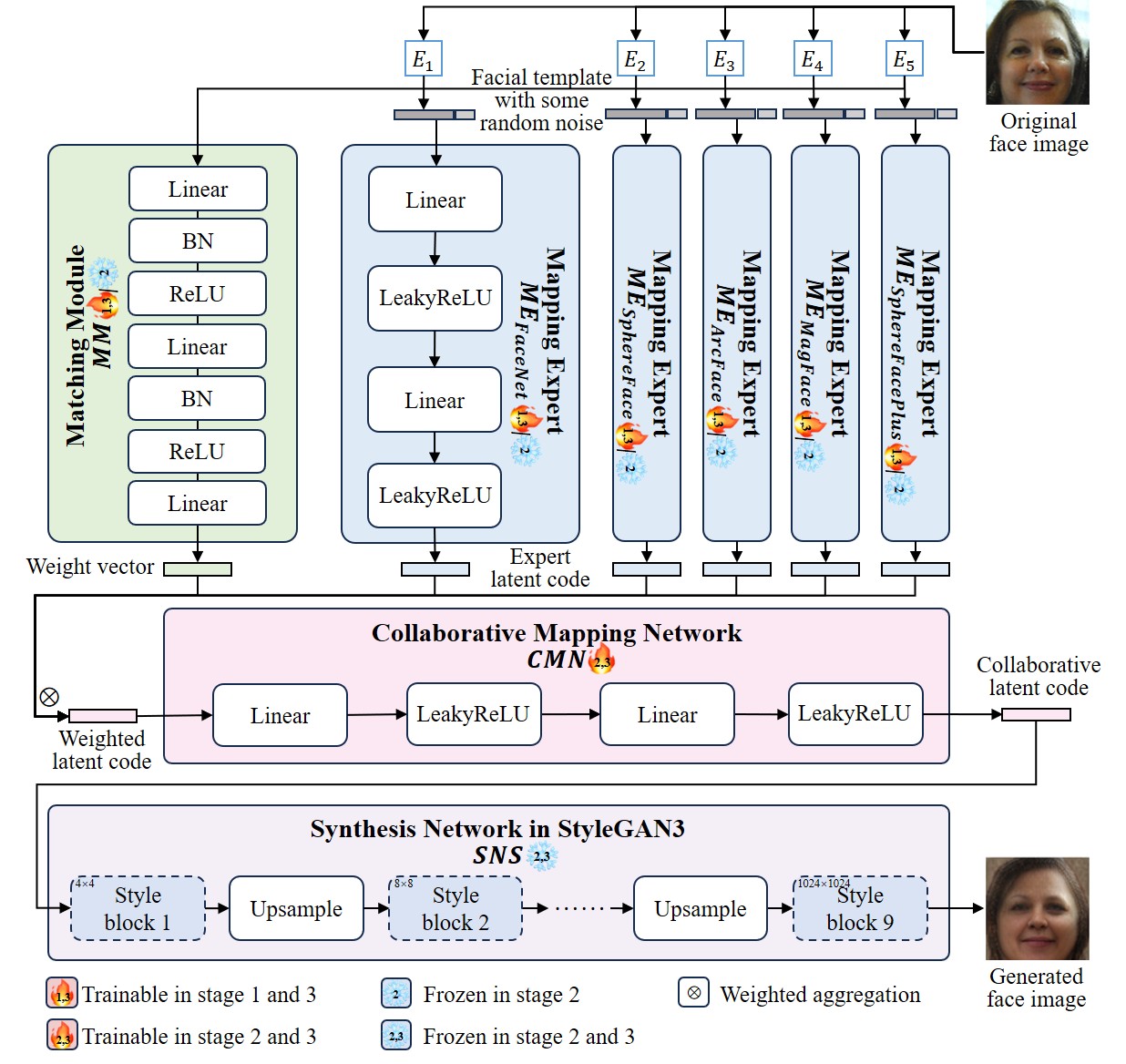} 
    \caption{Overall architecture of GMoE-FTI.}
    \vspace{-0.4cm}
    \label{fig:overall}
\end{figure}

Built upon the above taxonomy, we propose GMoE-FTI, a universal FTI framework built on genuine MoE design. Unlike prior FTI paradigms that only target a single FR extractor, our framework covers all five mainstream categories of facial template extractors. The overall architecture of our method is illustrated in Fig. \ref {fig:overall}. The core modules, loss functions, and training strategy are elaborated as follows.

\subsection{Core Modules}

%GMoE-FTI consists of five Mapping Experts (MEs), a matching module (MM), a Collaborative Mapping Network (CMN), and a pre-trained Synthesis Network of StyleGAN3 (SNS). Specifically, the MEs perform template-to-latent mapping for five typical facial template extractors. The MM module produces weights, and the CMN calculates the weighted code and optimizes it for SNS to reconstruction. To support arbitrary dimension of input, we standardize all templates to a fixed dimension of 1024 via zero-padding for vectors shorter than 1024 dimensions. The four core components are detailed below:

GMoE-FTI comprises five Mapping Experts (\textbf{ME}s), a Matching Module (\textbf{MM}), a Collaborative Mapping Network (\textbf{CMN}), and a pre-trained StyleGAN3 Synthesis Network (\textbf{SNS}). Specifically, each \textbf{ME} conducts template-to-latent mapping tailored to one of the five mainstream facial template extractors. The \textbf{MM} outputs expert weight coefficients, while the \textbf{CMN} aggregates weighted latent codes and refines them for subsequent facial reconstruction via the \textbf{SNS}. To accommodate input templates with arbitrary original dimensions, we uniformly standardize all templates to a fixed 1024-dimensional space: vectors with fewer than 1024 dimensions are padded with zeros. We elaborate on these four core components in the following subsections.

%\textbf{Mapping Expert (ME)}. We design five \textbf{ME} to learn template-to-latent mappings for five facial template extractors (FaceNet, SphereFace, ArcFace, MagFace, and SphereFacePlus) from five mainstream categories, respectively. All \textbf{ME}s have the same structure  with StyleGAN3-aligned mapping network, composing of two fully-connected layers and two LeakyReLU activations. The input consists of a 1024-dimensional template concatenated with 8-dimensional random noise, outputting a 512-dimensional expert code.

\textbf{Mapping Expert (ME)}. We design five \textbf{ME} modules to learn template-to-latent mappings corresponding to five facial template extractors (FaceNet, SphereFace, ArcFace, MagFace, and SphereFacePlus) from five mainstream categories, respectively. All \textbf{ME}s share an identical architecture derived from the StyleGAN3-aligned mapping network, which consists of two fully-connected layers followed by two LeakyReLU activation layers. Each \textbf{ME} module takes a 1024-dimensional facial template concatenated with 8-dimensional random noise as input and outputs a 512-dimensional expert latent code.

%\textbf{Matching Module (MM)}. The \textbf{MM} serves as the adaptive matching core of the entire framework. Functionally, it is a five-class classifier stacked with three linear layers, two batch normalization layers and two activation layers. Taking 1024-dimensional templates as input, it produces a 5-dimensional weight vector, each of which indicates the posterior probability that the input facial template comes from FaceNet, SphereFace, ArcFace, MagFace, and SphereFace+, respectively. The obtained weight vector is then used to aggregate the output of five \textbf{ME}s.

\textbf{Matching Module (MM)}. It acts as the adaptive matching core of our overall framework. Functionally, it is a five-category classifier stacked with three fully connected layers, two batch normalization layers, and two activation layers. Taking 1024-dimensional facial templates as input, it outputs a 5-dimensional weight vector, where each entry corresponds to the posterior probability that the input template originates from FaceNet, SphereFace, ArcFace, MagFace, and SphereFace+, respectively. The resulting weight vector is subsequently adopted to aggregate the outputs of five \textbf{ME}s.

%\textbf{Matching Module (MM)}. The MM serves as the adaptive matching core of the entire framework. It computes matching weights for feature streams output by distinct expert branches based on distribution differences between input templates and pre-defined template priors. Functionally, it is a five-class classifier stacked with three linear layers, two batch normalization layers and two activation layers. Taking 1024-dimensional templates as input, it produces a 5-dimensional weight vector, where each dimension represents the posterior probability that the input template comes from FaceNet, SphereFace, ArcFace, MagFace, and SphereFace+, respectively. The obtained weight vector enables weighted aggregation of the five expert latent codes to formulate a weighted code for subsequent collaborative mapping inversion.

%\textbf{Collaborative Mapping Network (CMN)}. \textbf{CMN} takes the weighted code as input and passes it through two fully-connected layers with LeakyReLU activations. It generates a code (termed as `collaborative code') that compatible to the synthesis network of StyleGAN3.

\textbf{Collaborative Mapping Network (CMN)}. It receives weighted latent codes as input and processes them via two fully-connected layers equipped with LeakyReLU activation functions. It outputs a so-called `collaborative latent code' that is compatible with the StyleGAN3 synthesis network.

%to achieve deep complementary fusion of multi-target identity features. The final output is a unified latent code (i.e., collaborative code) optimized across all expert branches.

%\textbf{Synthesis Network of StyleGAN3 (SNS)}. GMoE-FTI leverages a pre-trained SNS \cite{karras2021alias} as the final image generation backbone. The collaborative code is fed into the SNS pipeline, ultimately outputting identity-consistent face images with a resolution of $1024\times1024$.

\textbf{Synthesis Network in StyleGAN3 (SNS)}. GMoE-FTI adopts a pre-trained StyleGAN3 synthesis backbone \cite{karras2021alias} for final face reconstruction. The collaborative latent code is fed into the \textbf{SNS} pipeline, which ultimately outputs identity-consistent face images at a resolution of \(1024\times1024\).

\subsection{Loss Functions}

GMoE-FTI desiges four complementary loss functions for model training. The total loss is calculated as follows:
 \begin{equation}
 \mathcal{L}_{total} = \lambda_{w} \mathcal{L}_{w} + \lambda_{tem} \mathcal{L}_{tem} + \lambda_{pix} \mathcal{L}_{pix} + \lambda_{att} \mathcal{L}_{att},
 \end{equation}

 \noindent where $\lambda_{w}=\lambda_{tem}=\lambda_{pix}=1$, and $\lambda_{att}=10$. These hyperparameters balance the magnitude of individual loss terms, ensuring stable and equal optimization of all training objectives.

\begin{figure}[htbp]
    \centering
    \includegraphics[width=\linewidth]{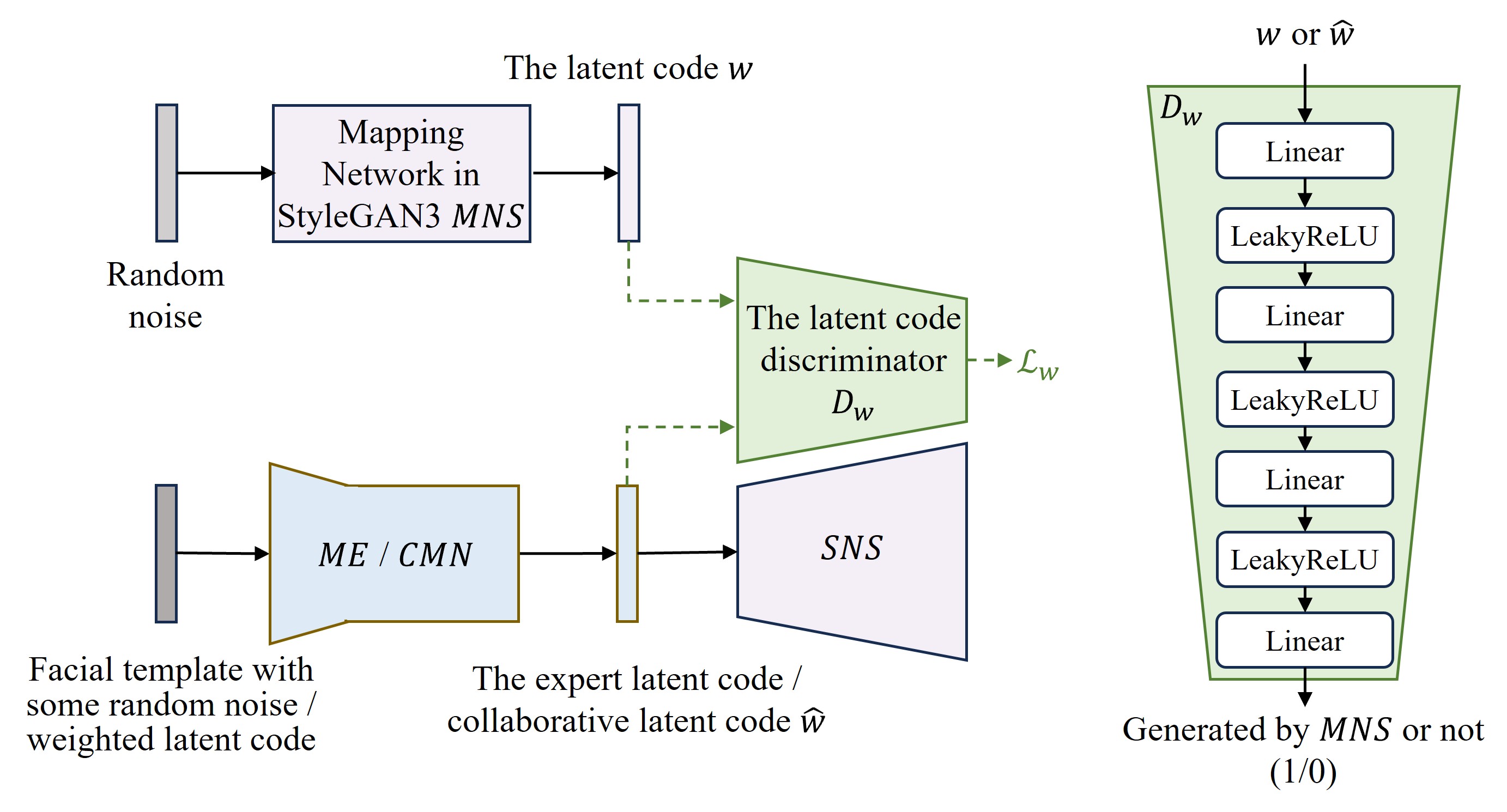} 
    \caption{The latent code loss and architecture of discriminator $D_w$.}
    \vspace{-0.4cm}
    \label{fig:dis}
\end{figure}

\paragraph{Latent code loss.} As shown in Fig. \ref{fig:dis}, we train an independent latent discriminator $D_w$ as a latent‑space adversarial regularizer to regularize latent codes output by individual \textbf{ME}s and unified latent codes from \textbf{CMN}, constraining generated latent codes $\hat{w}$ to match the latent distribution of StyleGAN3. The discriminative loss is formulated as:

\begin{equation}
    \mathcal{L}_w = \frac{1}{n} \sum_{i=1}^{n} D_w(\hat{w}_i),
\end{equation}

\noindent where $D_w(\hat{w}) \in [0,1]$, with the lower score for real and the higher for generated, and $n$ denotes the batch size.

\paragraph{Template loss.} It uses facial templates as strong identity supervision to constrain the templates extracted from reconstructed images to be similar with the ground-truth templates. We combine element-wise mean squared error (MSE) and cosine similarity loss on templates to lock core identity information strictly.
\begin{equation}
    \mathcal{L}_{tem} = \frac{1}{\|t\|} \sum_{i=1}^{\|t\|} (t_i - \hat{t}_i)^2 + 1 - \frac{t_i \times \hat{t}_i}{\|t_i\| \cdot \|\hat{t}_i\|},
\end{equation}

\noindent where $t_i$ and $\hat{t}_i$ denote the $i$-th element of the facial templates $t = \mathcal{E}(x)$ and $\hat{t} = \mathcal{E}(\hat{x})$, respectively. $\mathcal{E}(\cdot)$ denotes one of the five extractors (FaceNet, SphereFace, ArcFace, MagFace, SphereFace+).

\paragraph{Pixel loss.} It provides pixel-wise supervision which reduces per-pixel RGB discrepancies between reconstructed and original faces:
\begin{equation}
    \mathcal{L}_{pix} = \frac{1}{n} \sum_{i=1}^{n} (x_i - \hat{x}_i)^2,
\end{equation}

\noindent where $x_i$ and $\hat{x}_i$ denote the $i$-th pixel of the original face images $x$ and the generated one $\hat{x}$, respectively, and $n$ represents the pixel number.

\paragraph{Attribute loss.} To retain fine-grained facial attribute information during FTI, we introduce attribute loss for supervision. We employ a facial attribute classifier $\mathcal{A}(\cdot)$ pre-trained on the CelebA dataset\footnote{\url{https://github.com/Hawaii0821/FaceAttr-Analysis}}. The classifier outputs a 40-dimensional attribute vector, and we compute L1 distance between attribute vectors of original and reconstructed faces:
\begin{equation}
    \mathcal{L}_{att} = \frac{\sum_{i=1}^{40} |\mathcal{A}(x)_i - \mathcal{A}(\hat{x})_i|}{40},
\end{equation}

\noindent where $\mathcal{A}(x)_i$ and $\mathcal{A}(\hat{x})_i$ denote the $i$-th attribute of original and reconstructed images.

\subsection{Training Strategy}

We design a progressive three-stage training pipeline to ensuring stable model convergence and optimized inversion performance. 

\noindent \textbf{Stage 1: training MM and MEs independently}. The \textbf{MM} is pre-trained as a five-class classifier using softmax cross-entropy loss. We separately train five \textbf{ME}s. Each \textbf{ME} takes one type of facial templates concatenated with 8-dimensional random noise as input, and is optimized with all four loss functions defined above. This stage fully excavates independent mapping capacity for each \textbf{ME}.

\noindent \textbf{Stage 2: training CMN }. We freeze the parameters of \textbf{MM} and \textbf{ME}s trained in stage 1, alongside the pre-trained \textbf{SNS}. The \textbf{CMN} is trained with all four losses in this stage.

%. With expert latent codes aggregated into a weighted code via the weight vector output from the MM, the CMN optimizes this code to generate preliminary reconstructed face images.

\textbf{Stage 3: joint fine-tuning}. All models include \textbf{MM}, \textbf{ME}s, and \textbf{CMN}, except the pre-trained \textbf{SNS}, are fine-tuned with all four losses.

%are retained for joint supervision, further refining overall inversion performance and cross-distribution generalization while preserving the specialized capabilities of each pre-trained module.

\section{Experiments} 

\subsection{Experimental Setup} 

%Here we introduce experimental environment, datasets, and hyperparameters.

\textbf{Experimental environment}. All experiments are implemented on PyTorch with a single 24GB NVIDIA RTX 4090 GPU. The Adam optimizer is used for all training stages with $\beta_{1}$=0.5, $\beta_{2}$=0.9. All reconstructions maintain a fixed resolution of $1024 \times 1024$.

\noindent \textbf{Datasets}. We adopt two standard benchmarks with separated training and test splits for valid evaluation: FFHQ \cite{karras2019style} and LFW \cite{learned2016labeled}. Specifically, FFHQ (Flickr-Faces High-Quality) provides 70,000 unlabeled high-quality facial images for training. LFW (Labeled Faces in the Wild) serves as test set, containing 13,233 faces of 5,749 unique identities with diverse pose, facial expression, ethnicity, illumination, etc. 

\noindent \textbf{Training hyperparameters}. GMoE-FTI adopts the three-stage training strategy detailed above, with the specific hyperparameter settings as follows: 1) Stage 1 trains GMM to 100\% classification accuracy. All MEs are trained for 20 epochs with learning rate 0.0002 and batch size 32. 2) Stage 2 freezes all Stage 1 weights and SNS, only training CMN for 20 epochs with learning rate 0.00005 and batch size 1. 3) Stage 3 unlocks all modules of GMoE-FTI (except the frozen SNS) for global joint fine-tuning over 2 epochs, with learning rate 0.0001 and batch size 1.

\subsection{Evaluation Metrics}

\noindent \textbf{Verification precision}. Following the BLUFR protocol \cite{liao2014benchmark}, we report the True Accept Rate (TAR) at two fixed False Accept Rate (FAR) thresholds: 1\% and 0.1\%, to measure the machine authentication performance of reconstructed faces. Two types of TAR are defined for our task: 
\begin{itemize}
    \item Type-I TAR: It quantifies matching accuracy between reconstructed faces and their corresponding ground-truth original images.
    \item Type-II TAR: It evaluates matching performance on other samples belonging to the same identity, which reflects the cross-image generalization ability of FTI.

\end{itemize}

\noindent \textbf{Facial Attribute Pixel Deviation (FAPD)}: FAPD calculates pixel-wise errors within key facial foreground regions (eyebrows, eyes, nose, mouth). Facial landmark detection \cite{kazemi2014one} is applied to crop and align these regions. Lower FAPD indicates tighter pixel-level consistency of facial attributes between reconstructed and original images.

\noindent \textbf{Facial Attribute Perceptual Consistency (FAPC)}: FAPC measures the semantic feature distance of aligned facial foreground patches extracted from intermediate layers of pre-trained VGG-16 \cite{zhang2018unreasonable}. Smaller FAPC means higher human-perceived similarity of facial attributes.

\subsection{Comparative Experiments}

We perform comparative experiments with four representative SOTA FTI methods: Adv-StyleGAN3 (NeurIPS 2023) \cite{otroshi2023face}, MSE-StyleGAN3 (IJCB 2023) \cite{shahreza2023inversion}, DSCasConv (TIFS 2024) \cite{shahreza2024vulnerability}, and CLIP-FTI (AAAI 2026) \cite{dai2026clipfti}. Adv-StyleGAN3 and MSE-StyleGAN3 realize template-to-latent inversion via adversarial training and MSE constraints respectively. DSCasConv builds the inverter with stacked multi-scale deconvolution blocks, dense skip connections and parallel multi-kernel convolutions. CLIP-FTI introduces an extra CLIP semantic module to inject attribute-aware features into latent inversion.

\noindent \textbf{Quantitative results}. Tables \ref{tab:q_arc} to \ref{tab:q_ada} present comprehensive quantitative comparisons across diverse face template types. As presented in Table \ref{tab:q_arc}, GMoE-FTI achieves slightly degraded yet still favorable performance against SOTA baselines for inversion on in-training ArcFace templates. Tables \ref{tab:q_fac}, \ref{tab:q_sph}, \ref{tab:q_mag}, and \ref{tab:q_sphp} correspond to FaceNet, SphereFace, MagFace, and SphereFacePlus templates. These feature distributions are unseen domains for all compared SOTA baselines but belong to the training distribution of our GMoE-FTI. Tables \ref{tab:q_cur}, \ref{tab:q_ela}, \ref{tab:q_sph2}, \ref{tab:q_sphr}, \ref{tab:q_tra}, and \ref{tab:q_ada} further provide results on fully unseen templates including CosFace, CurricularFace, ElasticFace, SphereFace2, SphereFace-R, and TransFace.  When tackling these unseen FTI tasks, our GMoE-FTI delivers a 40\% average TAR gain, cutting average FAPD and FAPC by 2\% respectively. Our method surpasses all baseline competitors in identity consistency. These quantitative results verify that GMoE-FTI can effectively handle cross-feature-domain inversion tasks. 

\begin{table}[!htbp]
\centering
\scriptsize
\caption{Quantitative comparison with SOTA FTI methods on in-training ArcFace templates from LFW.}
\label{tab:q_arc}
\begin{tabular}{ccccccc}
\toprule
\multirow{2}{*}{Inverters} & \multicolumn{2}{c}{TAR at 1\% FAR} & \multicolumn{2}{c}{TAR at 0.1\% FAR} & \multirow{2}{*}{FAPD} & \multirow{2}{*}{FAPC} \\
\cmidrule(lr){2-3} \cmidrule(lr){4-5}
 & Type-I & Type-II & Type-I & Type-II & & \\
\midrule
DSCasConv & 0.9998 & 0.9908 & 0.9998 & 0.9836 & 0.0417 & 0.2555 \\
MSE-StyleGAN3 & 0.9992 & 0.8800 & 0.9879 & 0.6602 & 0.0569 & 0.2304 \\
Adv-StyleGAN3 & 1.0000 & 0.9294 & 0.9988 & 0.6980 & 0.0665 & 0.2423 \\
CLIP-FTI & 0.9998 & 0.8915 & 0.9922 & 0.6276 & 0.0514 & 0.2359 \\
\midrule
GMoE-FTI & 0.9982 & 0.7605 & 0.9915 & 0.5720 & 0.0517 & 0.2247 \\
\bottomrule
\end{tabular}
\end{table}

\begin{table}[!htbp]
\centering
\scriptsize
\caption{Quantitative comparison with SOTA FTI methods on FaceNet templates from LFW.}
\label{tab:q_fac}
\begin{tabular}{ccccccc}
\toprule
\multirow{2}{*}{Inverters} & \multicolumn{2}{c}{TAR at 1\% FAR} & \multicolumn{2}{c}{TAR at 0.1\% FAR} & \multirow{2}{*}{FAPD} & \multirow{2}{*}{FAPC} \\
\cmidrule(lr){2-3} \cmidrule(lr){4-5}
 & Type-I & Type-II & Type-I & Type-II & & \\
\midrule
DSCasConv & 0.0100 & 0.0023 & 0.0006 & 0.0001 & 0.0726 & 0.3108 \\
MSE-StyleGAN3 & 0.0100 & 0.0051 & 0.0011 & 0.0002 & 0.0415 & 0.2348 \\
Adv-StyleGAN3 & 0.0094 & 0.0027 & 0.0010 & 0.0001 & 0.0461 & 0.2267 \\
CLIP-FTI & 0.0097 & 0.0026 & 0.0008 & 0.0001 & 0.0390 & 0.2288 \\
\midrule
GMoE-FTI & \textbf{0.9974} & \textbf{0.7591} & \textbf{0.9907} & \textbf{0.5704} & \textbf{0.0519} & \textbf{0.2251} \\
\bottomrule
\end{tabular}
\end{table}

\begin{table}[!htbp]
\centering
\scriptsize
\caption{Quantitative comparison with SOTA FTI methods on SphereFace templates from LFW.}
\label{tab:q_sph}
\begin{tabular}{ccccccc}
\toprule
\multirow{2}{*}{Inverters} & \multicolumn{2}{c}{TAR at 1\% FAR} & \multicolumn{2}{c}{TAR at 0.1\% FAR} & \multirow{2}{*}{FAPD} & \multirow{2}{*}{FAPC} \\
\cmidrule(lr){2-3} \cmidrule(lr){4-5}
 & Type-I & Type-II & Type-I & Type-II & & \\
\midrule
DSCasConv & 0.0106 & 0.0030 & 0.0013 & 0.0001 & 0.0725 & 0.3109 \\
MSE-StyleGAN3 & 0.0100 & 0.0028 & 0.0008 & 0.0004 & 0.0415 & 0.2344 \\
Adv-StyleGAN3 & 0.0102 & 0.0033 & 0.0010 & 0.0001 & 0.0461 & 0.2265 \\
CLIP-FTI & 0.0092 & 0.0068 & 0.0008 & 0.0001 & 0.0389 & 0.2288 \\
\midrule
GMoE-FTI & \textbf{0.9706} & \textbf{0.6832} & \textbf{0.9528} & \textbf{0.4813} & \textbf{0.0568} & \textbf{0.2416} \\
\bottomrule
\end{tabular}
\end{table}

\begin{table}[!htbp]
\centering
\scriptsize
\caption{Quantitative comparison with SOTA FTI methods on MagFace templates from LFW.}
\label{tab:q_mag}
\begin{tabular}{ccccccc}
\toprule
\multirow{2}{*}{Inverters} & \multicolumn{2}{c}{TAR at 1\% FAR} & \multicolumn{2}{c}{TAR at 0.1\% FAR} & \multirow{2}{*}{FAPD} & \multirow{2}{*}{FAPC} \\
\cmidrule(lr){2-3} \cmidrule(lr){4-5}
 & Type-I & Type-II & Type-I & Type-II & & \\
\midrule
DSCasConv & 0.0088 & 0.0037 & 0.0011 & 0.0001 & 0.0392 & 0.2389 \\
MSE-StyleGAN3 & 0.0073 & 0.0031 & 0.0005 & 0.0001 & 0.0432 & 0.2373 \\
Adv-StyleGAN3 & 0.0075 & 0.0022 & 0.0007 & 0.0002 & 0.0480 & 0.2320 \\
CLIP-FTI & 0.0074 & 0.0022 & 0.0007 & 0.0001 & 0.0411 & 0.2343 \\
\midrule
GMoE-FTI & \textbf{0.9248} & \textbf{0.6125} & \textbf{0.8819} & \textbf{0.4147} & \textbf{0.0635} & \textbf{0.2629} \\
\bottomrule
\end{tabular}
\end{table}

\begin{table}[!htbp]
\centering
\scriptsize
\caption{Quantitative comparison with SOTA FTI methods on SphereFacePlus templates from LFW.}
\label{tab:q_sphp}
\begin{tabular}{ccccccc}
\toprule
\multirow{2}{*}{Inverters} & \multicolumn{2}{c}{TAR at 1\% FAR} & \multicolumn{2}{c}{TAR at 0.1\% FAR} & \multirow{2}{*}{FAPD} & \multirow{2}{*}{FAPC} \\
\cmidrule(lr){2-3} \cmidrule(lr){4-5}
 & Type-I & Type-II & Type-I & Type-II & & \\
\midrule
DSCasConv & 0.0085 & 0.0070 & 0.0010 & 0.0002 & 0.0725 & 0.31089 \\
MSE-StyleGAN3 & 0.0098 & 0.0053 & 0.0013 & 0.0005 & 0.0415 & 0.2344 \\
Adv-StyleGAN3 & 0.0098 & 0.0053 & 0.0013 & 0.0005 & 0.0461 & 0.2265 \\
CLIP-FTI &  0.0099 & 0.0019 & 0.0011 & 0.0001 & 0.0389 & 0.2288 \\
\midrule
GMoE-FTI & \textbf{0.9739} & \textbf{0.6908} & \textbf{0.9574} & \textbf{0.4897} & \textbf{0.0559} & \textbf{0.2376} \\
\bottomrule
\end{tabular}
\end{table}

\begin{table}[!htbp]
\centering
\scriptsize
\caption{Quantitative comparison with SOTA FTI methods on unseen CurricularFace templates from LFW.}
\label{tab:q_cur}
\begin{tabular}{ccccccc}
\toprule
\multirow{2}{*}{Inverters} & \multicolumn{2}{c}{TAR at 1\% FAR} & \multicolumn{2}{c}{TAR at 0.1\% FAR} & \multirow{2}{*}{FAPD} & \multirow{2}{*}{FAPC} \\
\cmidrule(lr){2-3} \cmidrule(lr){4-5}
 & Type-I & Type-II & Type-I & Type-II & & \\
\midrule
DSCasConv & 0.0099 & 0.0085 & 0.0013 & 0.0004 & 0.0417 & 0.2370 \\
MSE-StyleGAN3 & 0.0094 & 0.0032 & 0.0008 & 0.0001 & 0.0418 & 0.2371 \\
Adv-StyleGAN3 & 0.0088 & 0.0034 & 0.0009 & 0.0001 & 0.0461 & 0.2305 \\
CLIP-FTI & 0.0098 & 0.0258 & 0.0005 & 0.0017 & 0.0401 & 0.2332 \\
\midrule
GMoE-FTI & \textbf{0.3511} & \textbf{0.1960} & \textbf{0.2474} & \textbf{0.1545} & \textbf{0.0538} & \textbf{0.2317} \\
\bottomrule
\end{tabular}
\end{table}

\begin{table}[!htbp]
\centering
\scriptsize
\caption{Quantitative comparison with SOTA FTI methods on unseen ElasticFace templates from LFW.}
\label{tab:q_ela}
\begin{tabular}{ccccccc}
\toprule
\multirow{2}{*}{Inverters} & \multicolumn{2}{c}{TAR at 1\% FAR} & \multicolumn{2}{c}{TAR at 0.1\% FAR} & \multirow{2}{*}{FAPD} & \multirow{2}{*}{FAPC} \\
\cmidrule(lr){2-3} \cmidrule(lr){4-5}
 & Type-I & Type-II & Type-I & Type-II & & \\
\midrule
DSCasConv & 0.0115 & 0.0088 & 0.0014 & 0.0005 & 0.0500 & 0.2457 \\
MSE-StyleGAN3 & 0.0085 & 0.0026 & 0.0004 & 0.0001 & 0.0404 & 0.2357 \\
Adv-StyleGAN3 & 0.0088 & 0.0034 & 0.0009 & 0.0001 & 0.0473 & 0.2280 \\
CLIP-FTI & 0.0088 & 0.0041 & 0.0007 & 0.0002& 0.0394 & 0.2312 \\
\midrule
GMoE-FTI & \textbf{0.4703} & \textbf{0.2102} & \textbf{0.3637} & \textbf{0.1649} & \textbf{0.0521} & \textbf{0.2258} \\
\bottomrule
\end{tabular}
\end{table}

\begin{table}[!htbp]
\centering
\scriptsize
\caption{Quantitative comparison with SOTA FTI methods on unseen SphereFace2 templates from LFW.}
\label{tab:q_sph2}
\begin{tabular}{ccccccc}
\toprule
\multirow{2}{*}{Inverters} & \multicolumn{2}{c}{TAR at 1\% FAR} & \multicolumn{2}{c}{TAR at 0.1\% FAR} & \multirow{2}{*}{FAPD} & \multirow{2}{*}{FAPC} \\
\cmidrule(lr){2-3} \cmidrule(lr){4-5}
 & Type-I & Type-II & Type-I & Type-II & & \\
\midrule
DSCasConv & 0.0108 & 0.0012 & 0.0013 & 0.0001 & 0.0417 & 0.2370 \\
MSE-StyleGAN3 & 0.0091 & 0.0198 & 0.0008 & 0.0032 & 0.0413 & 0.2336 \\
Adv-StyleGAN3 & 0.0110 & 0.0434 & 0.0011 & 0.0048 & 0.0461 & 0.2261 \\
CLIP-FTI & 0.0116 & 0.0596 & 0.0013 & 0.0088 & 0.0390 & 0.2285 \\
\midrule
GMoE-FTI & \textbf{0.4465} & \textbf{0.1926} & \textbf{0.3419} & \textbf{0.1507} & \textbf{0.0546} & \textbf{0.2362} \\
\bottomrule
\end{tabular}
\end{table}

\begin{table}[!htbp]
\centering
\scriptsize
\caption{Quantitative comparison with SOTA FTI methods on unseen SphereFace-R templates from LFW.}
\label{tab:q_sphr}
\begin{tabular}{ccccccc}
\toprule
\multirow{2}{*}{Inverters} & \multicolumn{2}{c}{TAR at 1\% FAR} & \multicolumn{2}{c}{TAR at 0.1\% FAR} & \multirow{2}{*}{FAPD} & \multirow{2}{*}{FAPC} \\
\cmidrule(lr){2-3} \cmidrule(lr){4-5}
 & Type-I & Type-II & Type-I & Type-II & & \\
\midrule
DSCasConv & 0.0104 & 0.0090 & 0.0010 & 0.0001 & 0.0726 & 0.3109 \\
MSE-StyleGAN3 & 0.0094 & 0.0008 & 0.0004 & 0.0041 & 0.0415 & 0.2344 \\
Adv-StyleGAN3 & 0.0107 & 0.0018 & 0.0011 & 0.0001 & 0.0461 & 0.2265 \\
CLIP-FTI & 0.0091 & 0.0058 & 0.0008 & 0.0001 & 0.0389 & 0.2290 \\
\midrule
GMoE-FTI & \textbf{0.4620} & \textbf{0.2034} & \textbf{0.3562} & \textbf{0.1600} & \textbf{0.0530} & \textbf{0.2294} \\
\bottomrule
\end{tabular}
\end{table}

\begin{table}[!htbp]
\centering
\scriptsize
\caption{Quantitative comparison with SOTA FTI methods on unseen TransFace templates from LFW.}
\label{tab:q_tra}
\begin{tabular}{ccccccc}
\toprule
\multirow{2}{*}{Inverters} & \multicolumn{2}{c}{TAR at 1\% FAR} & \multicolumn{2}{c}{TAR at 0.1\% FAR} & \multirow{2}{*}{FAPD} & \multirow{2}{*}{FAPC} \\
\cmidrule(lr){2-3} \cmidrule(lr){4-5}
 & Type-I & Type-II & Type-I & Type-II & & \\
\midrule
DSCasConv & 0.0095 & 0.0014 & 0.0011 & 0.0002 & 0.0725 & 0.3105 \\
MSE-StyleGAN3 & 0.0091 & 0.0176 & 0.0010 & 0.0001 & 0.0409 & 0.2358 \\
Adv-StyleGAN3 & 0.0138 & 0.0115 & 0.0019 & 0.0012 & 0.0464 & 0.2266 \\
CLIP-FTI & 0.0097 & 0.0855 & 0.0006 & 0.0064 & 0.0391 & 0.2287 \\
\midrule
GMoE-FTI & \textbf{0.4393} & \textbf{0.1872} & \textbf{0.3356} & \textbf{0.1463} & \textbf{0.0554} & \textbf{0.2408} \\
\bottomrule
\end{tabular}
\end{table}

\begin{table}[!htbp]
\centering
\scriptsize
\caption{Quantitative comparison with SOTA FTI methods on unseen AdaFace templates from LFW.}
\label{tab:q_ada}
\begin{tabular}{ccccccc}
\toprule
\multirow{2}{*}{Inverters} & \multicolumn{2}{c}{TAR at 1\% FAR} & \multicolumn{2}{c}{TAR at 0.1\% FAR} & \multirow{2}{*}{FAPD} & \multirow{2}{*}{FAPC} \\
\cmidrule(lr){2-3} \cmidrule(lr){4-5}
 & Type-I & Type-II & Type-I & Type-II & & \\
\midrule
DSCasConv & 0.0104 & 0.0029 & 0.0008 & 0.0001 & 0.0725 & 0.3105 \\
MSE-StyleGAN3 & 0.0099 & 0.0018 & 0.0011 & 0.0001 & 0.0415 & 0.2344 \\
Adv-StyleGAN3 & 0.0110 & 0.0039 & 0.0012 & 0.0001 & 0.0461 & 0.2265 \\
CLIP-FTI & 0.0094 & 0.0089 & 0.0006 & 0.0004 & 0.0389 & 0.2289 \\
\midrule
GMoE-FTI & \textbf{0.4732} & \textbf{0.2121} & \textbf{0.3655} & \textbf{0.1675} & \textbf{0.0519} & \textbf{0.2248} \\
\bottomrule
\end{tabular}
\end{table}

\textbf{Quantitative results compared with independent Mapping Expert}.
We further conduct ablation experiments to evaluate the generalization performance of each standalone Mapping Expert. Tables \ref{tab:q_ME1} to \ref{tab:q_ME5} report the inversion results of individual Mapping Experts trained separately on FaceNet, SphereFace, ArcFace, MagFace, and SphereFacePlus, when fed with various types of face templates as inputs. The experimental results  indicate that every standalone expert suffers severe performance degradation when processing unseen out-of-distribution templates. Table \ref{tab:q_gmoe} lists the performance of our complete GMoE-FTI model, which validates that the cooperative fusion design of mixture-of-expert modules yields substantial performance improvements on cross-domain inversion tasks. Notably, CosFace produces 1024-dimensional facial templates, while existing SOTA methods only support 512-dimensional input features. For this reason, only our GMoE-FTI can report experimental results on CosFace templates.

\begin{table}[!htbp]
\centering
\scriptsize
\caption{Quantitative results of ME 1 on in-training FaceNet or other unseen templates from LFW.}
\label{tab:q_ME1}
\begin{tabular}{ccccccc}
\toprule
\multirow{2}{*}{Inverters} & \multicolumn{2}{c}{TAR at 1\% FAR} & \multicolumn{2}{c}{TAR at 0.1\% FAR} & \multirow{2}{*}{FAPD} & \multirow{2}{*}{FAPC} \\
\cmidrule(lr){2-3} \cmidrule(lr){4-5}
 & Type-I & Type-II & Type-I & Type-II & & \\
\midrule
FaceNet & 0.9998 & 0.9567 & 0.9917 & 0.7689 & 0.0394 & 0.2403 \\
SphereFace & 0.0082 & 0.0032 & 0.0006 & 0.0001 & 0.0368 & 0.2341 \\
ArcFace & 0.0078 & 0.0020 & 0.0009 & 0.0001 & 0.1061 & 0.3569 \\
MagFace & 0.0085 & 0.0034 & 0.0009 & 0.0002 & 0.1251 & 0.3853 \\
SphereFace+ & 0.0096 & 0.0016 & 0.0008 & 0.0002 & 0.0366 & 0.2337 \\
CosFace & 0.0129 & 0.0053 & 0.0018 & 0.0004 & 0.1416 & 0.4095 \\
CurricularFace & 0.0093 & 0.0223 & 0.0008 & 0.0012 & 0.1233 & 0.3769 \\
ElasticFace & 0.0104 & 0.0196 & 0.0010 & 0.0014 & 0.0950 & 0.3320 \\
SphereFace2 & 0.0085 & 0.0035 & 0.0008 & 0.0003 & 0.0388 & 0.2338 \\
SphereFace-R & 0.0116 & 0.0070 & 0.0008 & 0.0001 & 0.0363 & 0.2334 \\
TransFace & 0.0092 & 0.0052 & 0.0008 & 0.0002 & 0.0390 & 0.2363 \\
AdaFace & 0.0117 & 0.0044 & 0.0015 & 0.0002 & 0.0363 & 0.2340 \\
\bottomrule
\end{tabular}
\end{table}

\begin{table}[!htbp]
\centering
\scriptsize
\caption{Quantitative results of ME 2 on in-training SphereFace or other unseen templates from LFW.}
\label{tab:q_ME2}
\begin{tabular}{ccccccc}
\toprule
\multirow{2}{*}{Inverters} & \multicolumn{2}{c}{TAR at 1\% FAR} & \multicolumn{2}{c}{TAR at 0.1\% FAR} & \multirow{2}{*}{FAPD} & \multirow{2}{*}{FAPC} \\
\cmidrule(lr){2-3} \cmidrule(lr){4-5}
 & Type-I & Type-II & Type-I & Type-II & & \\
\midrule
FaceNet & 0.0060 & 0.0010 & 0.0002 & 0.0001 & 0.0349 & 0.2359 \\
SphereFace & 0.9922 & 0.8199 & 0.9473 & 0.5051 & 0.0400 & 0.2372 \\
ArcFace & 0.0073 & 0.0022 & 0.0008 & 0.0002 & 0.1052 & 0.3282 \\
MagFace & 0.0084 & 0.0049 & 0.0011 & 0.0004 & 0.1019 & 0.3194 \\
SphereFace+ & 0.0072 & 0.0034 & 0.0008 & 0.0001 & 0.0357 & 0.2356 \\
CosFace & 0.0102 & 0.0595 & 0.0011 & 0.0073 & 0.1004 & 0.3279 \\
CurricularFace & 0.0085 & 0.0057 & 0.0006 & 0.0002 & 0.0943 & 0.3093 \\
ElasticFace & 0.0091 & 0.0036 & 0.0005 & 0.0002 & 0.0920 & 0.3017 \\
SphereFace2 & 0.0106 & 0.0183 & 0.0013 & 0.0015 & 0.0335 & 0.2334 \\
SphereFace-R & 0.0072 & 0.0034 & 0.0008 & 0.0001 & 0.0340 & 0.2359 \\
TransFace & 0.0085 & 0.0177 & 0.0007 & 0.0001 & 0.0520 & 0.2473 \\
AdaFace & 0.0057 & 0.0021 & 0.0007 & 0.0001 & 0.0351 & 0.2363 \\
\bottomrule
\end{tabular}
\end{table}

\begin{table}[!htbp]
\centering
\scriptsize
\caption{Quantitative results of ME 3 on in-training ArcFace or other unseen templates from LFW.}
\label{tab:q_ME3}
\begin{tabular}{ccccccc}
\toprule
\multirow{2}{*}{Inverters} & \multicolumn{2}{c}{TAR at 1\% FAR} & \multicolumn{2}{c}{TAR at 0.1\% FAR} & \multirow{2}{*}{FAPD} & \multirow{2}{*}{FAPC} \\
\cmidrule(lr){2-3} \cmidrule(lr){4-5}
 & Type-I & Type-II & Type-I & Type-II & & \\
\midrule
FaceNet & 0.0094 & 0.0021 & 0.0005 & 0.0001 & 0.0482 & 0.2238 \\
SphereFace & 0.0121 & 0.0475 & 0.0014 & 0.0053 & 0.0481 & 0.2239 \\
ArcFace & 0.9999 & 0.8868 & 0.9982 & 0.6220 & 0.0444 & 0.2353 \\
MagFace & 0.0091 & 0.0030 & 0.0006 & 0.0001 & 0.0465 & 0.2310 \\
SphereFace+ & 0.0110 & 0.0042 & 0.0008 & 0.0001 & 0.0481 & 0.2239 \\
CosFace & 0.0138 & 0.0072 & 0.0023 & 0.0014 & 0.0494 & 0.2389 \\
CurricularFace & 0.0089 & 0.0069 & 0.0009 & 0.0003 & 0.0453 & 0.2293 \\
ElasticFace & 0.0098 & 0.0031 & 0.0008 & 0.0001 & 0.0460 & 0.2269 \\
SphereFace2 & 0.0106 & 0.0099 & 0.0008 & 0.0001 & 0.0481 & 0.2241 \\
SphereFace-R & 0.0106 & 0.0094 & 0.0009 & 0.0003 & 0.0482 & 0.2239 \\
TransFace & 0.0093 & 0.0198 & 0.0014 & 0.0008 & 0.0482 & 0.2240 \\
AdaFace & 0.0106 & 0.0055 & 0.0011 & 0.0002 & 0.0482 & 0.2240 \\
\bottomrule
\end{tabular}
\end{table}

\begin{table}[!htbp]
\centering
\scriptsize
\caption{Quantitative results of ME 4 on in-training MagFace or other unseen templates from LFW.}
\label{tab:q_ME4}
\begin{tabular}{ccccccc}
\toprule
\multirow{2}{*}{Inverters} & \multicolumn{2}{c}{TAR at 1\% FAR} & \multicolumn{2}{c}{TAR at 0.1\% FAR} & \multirow{2}{*}{FAPD} & \multirow{2}{*}{FAPC} \\
\cmidrule(lr){2-3} \cmidrule(lr){4-5}
 & Type-I & Type-II & Type-I & Type-II & & \\
\midrule
FaceNet & 0.0093 & 0.0023 & 0.0007 & 0.0001 & 0.0423 & 0.2372 \\
SphereFace & 0.0101 & 0.0017 & 0.0011 & 0.0002 & 0.0423 & 0.2373 \\
ArcFace & 0.0077 & 0.0034 & 0.0011 & 0.0002 & 0.0404 & 0.2391 \\
MagFace & 0.9971 & 0.8200 & 0.9598 & 0.4491 & 0.0437 & 0.2422 \\
SphereFace+ & 0.0082 & 0.0048 & 0.0010 & 0.0003 & 0.0423 & 0.2373 \\
CosFace & 0.0105 & 0.0021 & 0.0009 & 0.0001 & 0.0462 & 0.2414 \\
CurricularFace & 0.0082 & 0.0015 & 0.0013 & 0.0002 & 0.0399 & 0.2402 \\
ElasticFace & 0.0094 & 0.0090 & 0.0008 & 0.0002 & 0.0401 & 0.2387 \\
SphereFace2 & 0.0085 & 0.0098 & 0.0011 & 0.0005 & 0.0424 & 0.2372 \\
SphereFace-R & 0.0084 & 0.0099 & 0.0011 & 0.0004 & 0.0424 & 0.2373 \\
TransFace & 0.0108 & 0.0021 & 0.0011 & 0.0002 & 0.0422 & 02372. \\
AdaFace & 0.0089 & 0.0098 & 0.0009 & 0.0007 & 0.0423& 0.2372 \\
\bottomrule
\end{tabular}
\end{table}

\begin{table}[!htbp]
\centering
\scriptsize
\caption{Quantitative results of ME 5 on in-training SphereFace+ or other unseen templates from LFW.}
\label{tab:q_ME5}
\begin{tabular}{ccccccc}
\toprule
\multirow{2}{*}{Inverters} & \multicolumn{2}{c}{TAR at 1\% FAR} & \multicolumn{2}{c}{TAR at 0.1\% FAR} & \multirow{2}{*}{FAPD} & \multirow{2}{*}{FAPC} \\
\cmidrule(lr){2-3} \cmidrule(lr){4-5}
 & Type-I & Type-II & Type-I & Type-II & & \\
\midrule
FaceNet & 0.0062 & 0.0026 & 0.0003 & 0.0001 & 0.0343 & 0.2376 \\
SphereFace & 0.0196 & 0.0139 & 0.0031 & 0.0013 & 0.0353 & 0.2365 \\
ArcFace & 0.0118 & 0.0160 & 0.0008 & 0.0017 & 0.0930 & 0.3189 \\
MagFace & 0.0096 & 0.0064 & 0.0007 & 0.0002 & 0.0873 & 0.3070 \\
SphereFace+ & 0.9937 & 0.9569 & 0.9621 & 0.8209 & 0.0390 & 0.2375 \\
CosFace & 0.0113 & 0.0550 & 0.0012 & 0.0055 & 0.1000 & 0.3264 \\
CurricularFace & 0.0110 & 0.0043 & 0.0005 & 0.0002 & 0.0941 & 0.3203 \\
ElasticFace & 0.0086 & 0.0036 & 0.0007 & 0.0003 & 0.0912 & 0.3026 \\
SphereFace2 & 0.0113 & 0.0046 & 0.0018 & 0.0002 & 0.0377 & 0.2375 \\
SphereFace-R & 0.0076 & 0.0021 & 0.0009 & 0.0001 & 0.0354 & 0.2357 \\
TransFace & 0.0074 & 0.0029 & 0.0006 & 0.0002 & 0.0474 & 0.2445 \\
AdaFace & 0.0183 & 0.0767 & 0.0022 & 0.0111 & 0.348 & 0.2358 \\
\bottomrule
\end{tabular}
\end{table}

\begin{table}[!htbp]
\centering
\scriptsize
\caption{Quantitative results of GMoE-FTI on in-training or other unseen templates from LFW.}
\label{tab:q_gmoe}
\begin{tabular}{ccccccc}
\toprule
\multirow{2}{*}{Inverters} & \multicolumn{2}{c}{TAR at 1\% FAR} & \multicolumn{2}{c}{TAR at 0.1\% FAR} & \multirow{2}{*}{FAPD} & \multirow{2}{*}{FAPC} \\
\cmidrule(lr){2-3} \cmidrule(lr){4-5}
 & Type-I & Type-II & Type-I & Type-II & & \\
\midrule
FaceNet & 0.9974 & 0.7591 & 0.9907 & 0.5704 & 0.0519 & 0.2251 \\
SphereFace & 0.9706 & 0.6832 & 0.9528 & 0.4813 & 0.0568 & 0.2416 \\
ArcFace &  0.9982 & 0.7605 & 0.9915 & 0.5720 & 0.0517 & 0.2247 \\
MagFace & 0.9248 & 0.6125 & 0.8819 & 0.4147 & 0.0635 & 0.2629 \\
SphereFace+ & 0.9739 & 0.6908 & 0.9574 & 0.4897 & 0.0559 & 0.2376 \\
CurricularFace & 0.3511 & 0.1960 & 0.2474 & 0.1545 & 0.0538 & 0.2317 \\
ElasticFace & 0.4703 & 0.2102 & 0.3637 & 0.1649 & 0.0521 & 0.2258 \\
SphereFace2 & 0.4465 & 0.1926 & 0.3419 & 0.1507 & 0.0546 & 0.2362 \\
SphereFace-R & 0.4620 & 0.2034 & 0.3562 & 0.1600 & 0.0530 & 0.2294 \\
TransFace & 0.4393 & 0.1872 & 0.3356 & 0.1463 & 0.0554 & 0.2408 \\
AdaFace & 0.4732 & 0.2121 & 0.3655 & 0.1675 & 0.0519 & 0.2248 \\
CosFace & 0.3680 & 0.2073 & 0.2610 & 0.1631 & 0.0524 & 0.2265 \\
\bottomrule
\end{tabular}
\end{table}

\begin{figure*}[!htbp]
    \centering
    \includegraphics[width=\linewidth]{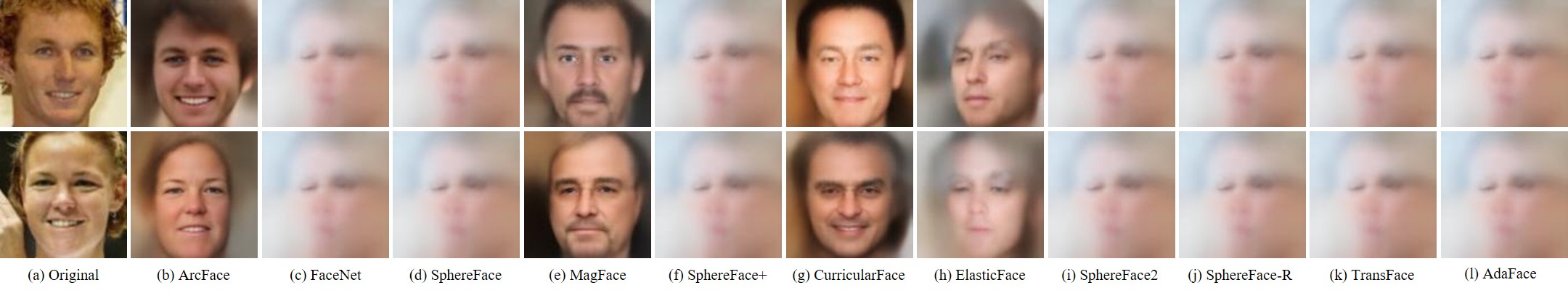}
    \vspace{-0.45cm}
    \caption{Qualitative  comparison against DSCasConv on different templates from LFW.}
    \label{fig:q_tifs}
\end{figure*}

\begin{figure*}[!htbp]
    \centering
    \includegraphics[width=\linewidth]{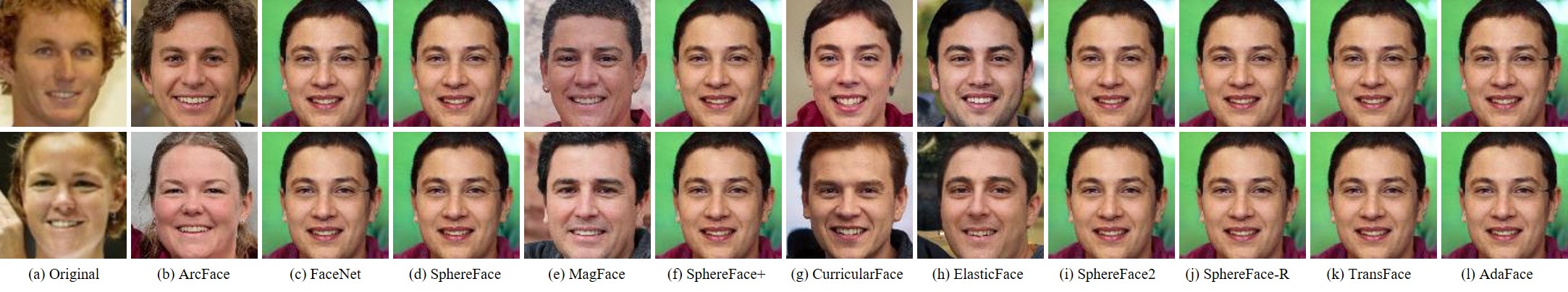}
    \vspace{-0.45cm}
    \caption{Qualitative  comparison against MSE-StyleGAN3 on different templates from LFW.}
    \label{fig:q_ijcb}
\end{figure*}

\begin{figure*}[!htbp]
    \centering
    \includegraphics[width=\linewidth]{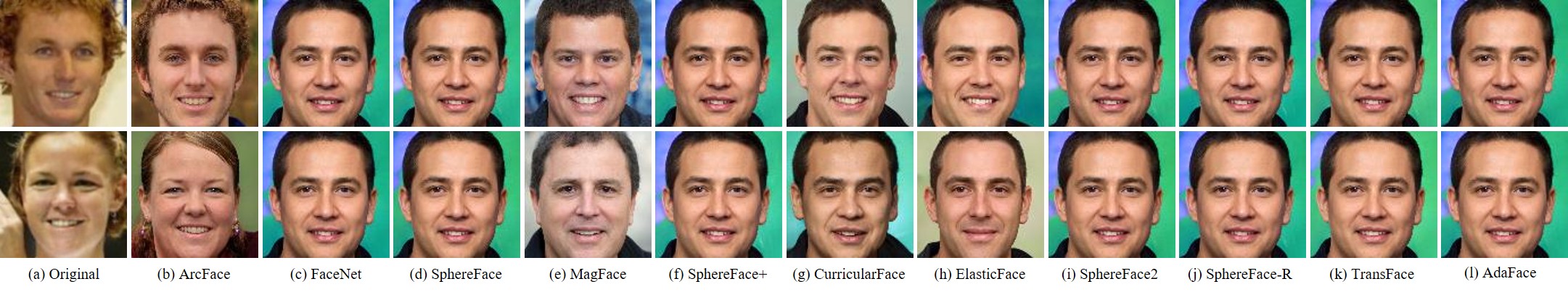}
    \vspace{-0.45cm}
    \caption{Qualitative  comparison against Adv-StyleGAN3 on different templates from LFW.}
    \label{fig:q_nips}
\end{figure*}

\begin{figure*}[!htbp]
    \centering
    \includegraphics[width=\linewidth]{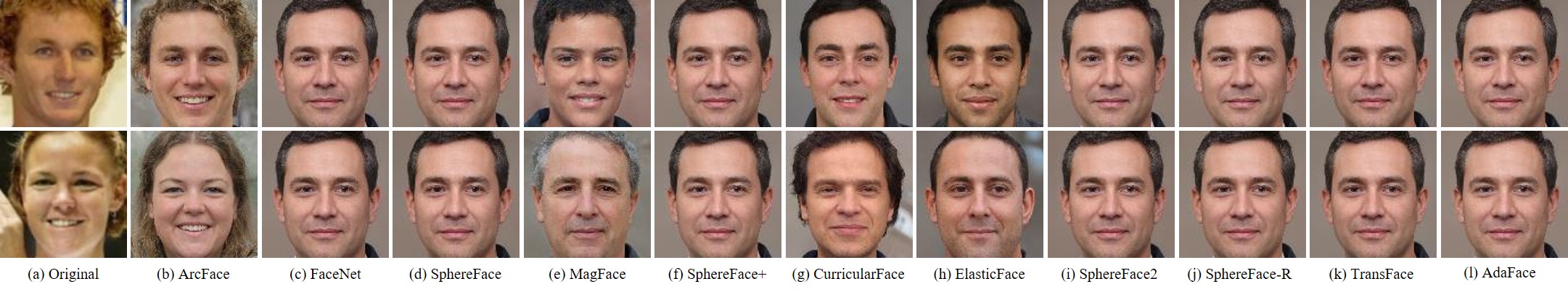}
    \vspace{-0.45cm}
    \caption{Qualitative  comparison against CLIP-FTI on different templates from LFW.}
    \label{fig:q_aaai}
\end{figure*}

\begin{figure*}[!htbp]
    \centering
    \includegraphics[width=\linewidth]{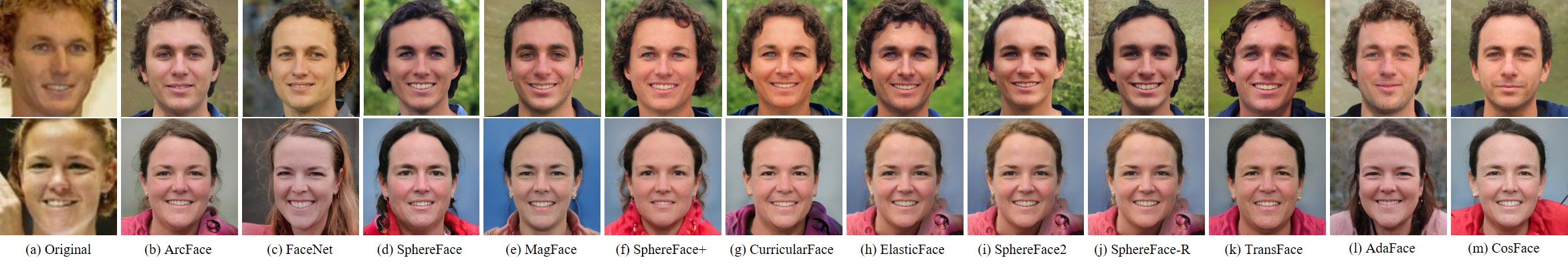}
    \vspace{-0.45cm}
    \caption{Qualitative  comparison against GMoE-FTI on different templates from LFW.}
    \label{fig:q_moe}
\end{figure*}

\noindent \textbf{Qualitative visualization}. Figs. \ref{fig:q_tifs} to \ref{fig:q_aaai} show that all SOTA baselines only achieve satisfying reconstructions on ArcFace templates. When applied to unseen FR templates, they suffer from prominent identity-inconsistency and severe facial distortion, including mismatched facial features and blurriness. As shown in Fig. \ref{fig:q_moe}, our GMoE-FTI produces clear, identity-consistent face images even on unseen templates. This stable cross-backbone performance proves that our GMoE-FTI is a universal FTI framework.

\subsection{Ablation Experiments}

%To verify the indispensability of each core component of GMoE-FTI, we conduct four groups of ablation experiments on the LFW dataset using unseen AdaFace templates as input. As summarized in Table \ref{tab:ablation}, the first four rows report the performance of four ablated model variants, and the fifth row presents results of the complete full-version GMoE-FTI. Figure 5 provides partial qualitative visualizations under the same unseen AdaFace template test setting from LFW. Experimental results reveal the necessity of each component as follows:

We conduct ablation studies on unseen AdaFace templates from LFW to validate the necessity of each core component. Table \ref{tab:ablation} lists four ablated variants and the complete full model. The key findings are summarized below:

%\begin{itemize}
    %\item Discarding the MM classification module disrupts adaptive gated routing between expert branches, resulting in mismatched cross-branch feature fusion and prominent identity loss in reconstructed faces.
    %\item Replacing our pre-trained heterogeneous experts with conventional load-balanced MoE severely degrades generalization on unseen template distributions, which verifies that independently pre-trained dedicated experts act as the core support for our universal inversion ability. 
    %\item Removing the self-attention unit inside CMN weakens the model’s capability to capture long-range semantic dependencies of facial structures, producing distorted local facial details and higher FAPD and FAPC scores. 
    %\item Eliminating the joint fine-tuning stage destroys end-to-end collaborative optimization across all modules; each submodule can no longer coordinate adaptively to fit unseen template distributions.
%\end{itemize}

\begin{itemize}

\item Replacing GMoE with conventional load-balanced MoE hurts generalization on unseen templates. Such MoE adopts static expert allocation and uniform load balancing and cannot adapt to heterogeneous template types, proving our task-specific pre-trained experts are vital for universal inversion.
\item Removing the \textbf{MM} adaptive weighting module forces equal averaging of all five experts’ latent features. Without template-aware weight assignment, cross-branch fusion collapses and severe identity drift emerges, accompanied by a significant TAR decline.remarkable drop in TAR.
\item Feeding the concatenated weight vector and five expert latent codes directly into the \textbf{CMN} module without weighted aggregation damages long-range facial semantic representation, which introduces severe local texture distortion and increases FAPD and FAPC.
\item Without joint fine-tuning, cross-module end-to-end coordination collapses, disabling adaptive fitting for diverse unseen template distributions and reducing overall reconstruction quality.
\end{itemize}

\begin{table}[ht]
\centering
\tiny
\begin{threeparttable}
\caption{Results of ablation experiments.}
\label{tab:ablation}
\begin{tabular}{m{0.3cm}m{0.3cm}m{0.2cm}m{0.2cm}m{0.6cm}m{0.65cm}m{0.6cm}m{0.65cm}m{0.4cm}m{0.4cm}}
\toprule
\multirow{2}{*}{GMoE} & \multirow{2}{*}{MM} & \multirow{2}{*}{W-G} & \multirow{2}{*}{S3} & \multicolumn{2}{c}{TAR at 1\% FAR} & \multicolumn{2}{c}{TAR at 0.1\% FAR} & \multirow{2}{*}{FAPD} & \multirow{2}{*}{FAPC} \\
\cmidrule(lr){5-6} \cmidrule(lr){7-8}
 & & & & Type-I & Type-II & Type-I & Type-II & & \\
\midrule
$\times$ & $\checkmark$ & $\checkmark$ & $\checkmark$ & 0.0109 & 0.0040 & 0.0018 & 0.0003 & 0.1176 & 0.4628 \\
$\checkmark$ & $\times$ & $\checkmark$ & $\checkmark$ & 0.0113 & 0.0128 & 0.0009 & 0.0002 & 0.0821 & 0.3677 \\
$\checkmark$ & $\checkmark$ & $\times$ & $\checkmark$ & 0.0104 & 0.0037 & 0.0006 & 0.0002 & 0.0822 & 0.3680 \\
$\checkmark$ & $\checkmark$ & $\checkmark$ & $\times$ & 0.4725 & 0.2014 & 0.3643 & 0.1599 & 0.0524 & 0.2447 \\
$\checkmark$ & $\checkmark$ & $\checkmark$ & $\checkmark$ & \textbf{0.4732} & \textbf{0.2121} & \textbf{0.3655} & \textbf{0.1675} & \textbf{0.0517} & \textbf{0.2248} \\
\bottomrule
\end{tabular}
\begin{tablenotes}[flushleft]
    \tiny 
    \item[†] Note: GMoE distinguishes our customized expert framework from existing load-balanced MoE. W-G denotes whether weighted aggregation is performed prior to feeding into the CMN module. S3 corresponds to the ablation setting that removes the full-module joint fine-tuning stage.
\end{tablenotes}
\end{threeparttable}
\end{table}

\section{Conclusion} 

To address the fixed feature domain bias and weak generalization of conventional FTI methods, this paper proposes GMoE-FTI, a universal FTI framework built upon a genuine MoE mechanism. Centered on five independently pre-trained expert branches, GMoE-FTI integrates adaptive matching and collaborative joint inference to realize universal identity-consistent reconstruction from five types of prevalent facial template extractors. Extensive quantitative and qualitative experiments verify that GMoE-FTI delivers significant performance gains on unseen template inversion tasks, outperforming SOTA models on TAR, FAPD, and FAPC. For future work, we will further optimize the number and allocation strategy of expert branches, boost inversion accuracy for facial templates, and expand the application scope of our framework to biometric privacy defense systems.

\bibliographystyle{IEEEtran}

\bibliography{resf}

\vfill

\end{document}